\documentclass[a4paper,10pt,twocolumn]{article}
\usepackage{times}
\usepackage{url}
\usepackage{graphicx}
\usepackage{multirow}
\usepackage{amsmath} 

\makeatletter
\def\section{\@startsection {section}{1}{\z@}{-3.5ex \@plus -1ex \@minus -.2ex}{2.3ex \@plus.2ex}{\normalsize\bf\centering}}
\makeatother 
\makeatletter
\def\@seccntformat#1{\csname the#1\endcsname.~}
\makeatother

\makeatletter
\def\subsection{\@startsection {subsection}{2}{\z@}{-3.25ex\@plus -1ex \@minus -.2ex}{1.5ex \@plus .2ex}{\normalsize\bf}}
\makeatother 

\makeatletter
\def\subsubsection{\@startsection {subsubsection}{3}{\z@}{-3.25ex\@plus -1ex \@minus -.2ex}{1.5ex \@plus .2ex}{\normalsize\bf}}
\makeatother 

\title{\Large \bf {
Accurate and Efficient Surface Reconstruction from Point Clouds 

via Geometry-Aware Local Adaptation
}}
\author{
Eito Ogawa\\
School of FSE, \\Waseda University
\and
Taiga Hayami\\
Graduate School of FSE, \\Waseda University
\and
Hiroshi Watanabe\\
Graduate School of FSE, \\Waseda University}

\date{\vspace{-9mm}} 

\begin{document}
\maketitle
\thispagestyle{empty} 

\section*{Abstract}
Point cloud surface reconstruction has improved in accuracy with advances in deep learning, enabling applications such as infrastructure inspection. Recent approaches that reconstruct from small local regions rather than entire point clouds have attracted attention for their strong generalization capability. However, prior work typically places local regions uniformly and keeps their size fixed, limiting adaptability to variations in geometric complexity. In this study, we propose a method that improves reconstruction accuracy and efficiency by adaptively modulating the spacing and size of local regions based on the curvature of the input point cloud.
\vspace{1mm}

{\noindent \bf Keywords:} Point cloud, mesh, unsigned distance field
\vspace{3mm}

\section{Introduction}
Surface reconstruction from point clouds is in increasing demand as the growing availability of LiDAR expands opportunities to generate 3D models from point clouds.
With the development of deep learning, numerous neural network based approaches have been proposed.
Among these, implicit neural representations\cite{deepsdf}\cite{onet} have gained significant influence as a framework that learns scalar fields assigning numerical values such as distance to each point in space.
Despite strong performance on watertight surfaces, their dependence on inside–outside definitions constrains the modeling of non-watertight geometries featuring edges and open boundaries.
To circumvent this constraint, unsigned distance fields (UDFs), which represent the shortest unsigned distance to the surface for each point in space, have attracted attention.
Since UDF does not assume interior-exterior classification, it enables reconstruction of not only non-watertight surfaces but also watertight surfaces.
LoSF-UDF \cite{losf}  achieves excellent reconstruction quality by dividing the target point cloud into small local regions for reconstruction.
However, local region sizes are fixed, which frequently results in the mixing of multiple surfaces or layers in high-curvature regions and insufficient point density within those regions.
These factors hinder the faithful capture of local geometric structure.

In this paper we propose a point cloud surface reconstruction method that adapts the size of each local region to geometric complexity. 
Complexity is quantified by curvature and the patch radius is adjusted accordingly. UDF query points are placed with variable resolution, starting from a uniform grid of $128^3$ and locally refined to $256^3$ in highly curved areas. Resampling is also conditioned on curvature. Together these components suppress layer mixing in complex regions, secure sufficient evidence in smooth regions, and reduce average computational cost while preserving stable UDF estimation and high-quality reconstruction.

\begin{figure}[t]
\centering
\includegraphics[width=\columnwidth]{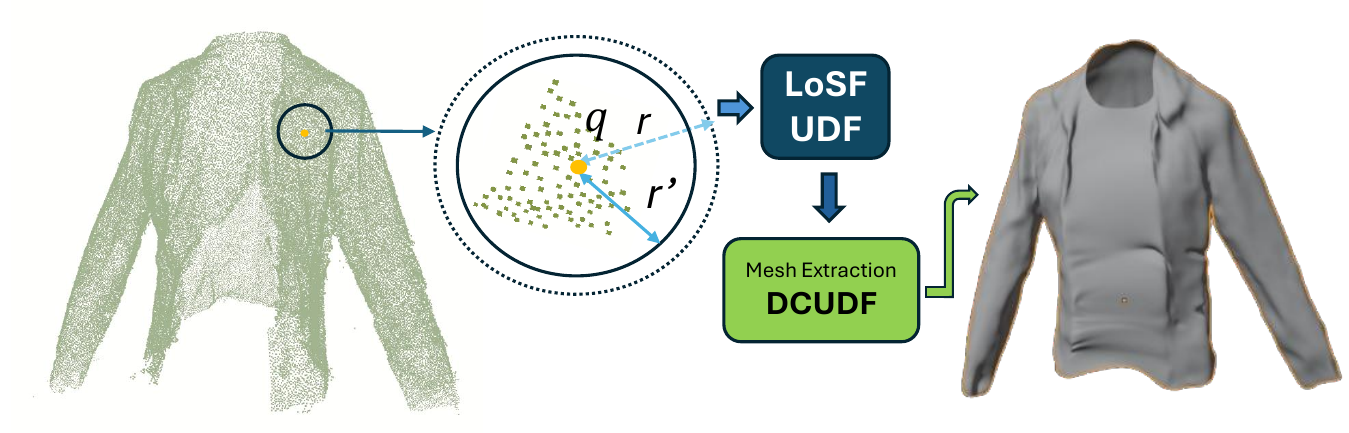}
\caption{Reconstruction pipeline. We extract a local patch assigned to each query point $q$ within
a radius $r'$ based on the curvature and obtain the UDF values.
Then we extract the mesh from the UDF field with DCUDF\cite{dcudf}.}
\label{fig:pipeline}
\end{figure}

\section{Related Work}
\subsection{Learning implicit surface representation}
Implicit surface representations have attracted significant attention for their ability to represent 3D shapes as continuous functions and flexibly select resolution during mesh extraction.
Representative methods within this framework include those based on Signed Distance Fields (SDFs) \cite{deepsdf} and occupancy fields\cite{onet}.
SDF represents the signed distance from any point in 3D space to the nearest object surface as a scalar value. 
This property enables rational control of loss function design and convergence criteria based on distance magnitudes, and naturally reproduces watertight surfaces even for complex geometries.
Occupancy fields assign to each spatial coordinate the probability of being inside the object, and the surface is recovered as the level set of this probability at a chosen threshold.
The required supervision signal consists only of interior-exterior labels, which are readily obtainable.
However, since these methods presuppose interior-exterior classification, they face difficulties in representing non-watertight surfaces.

\begin{figure}[t]
\centering
\includegraphics[width=\columnwidth]{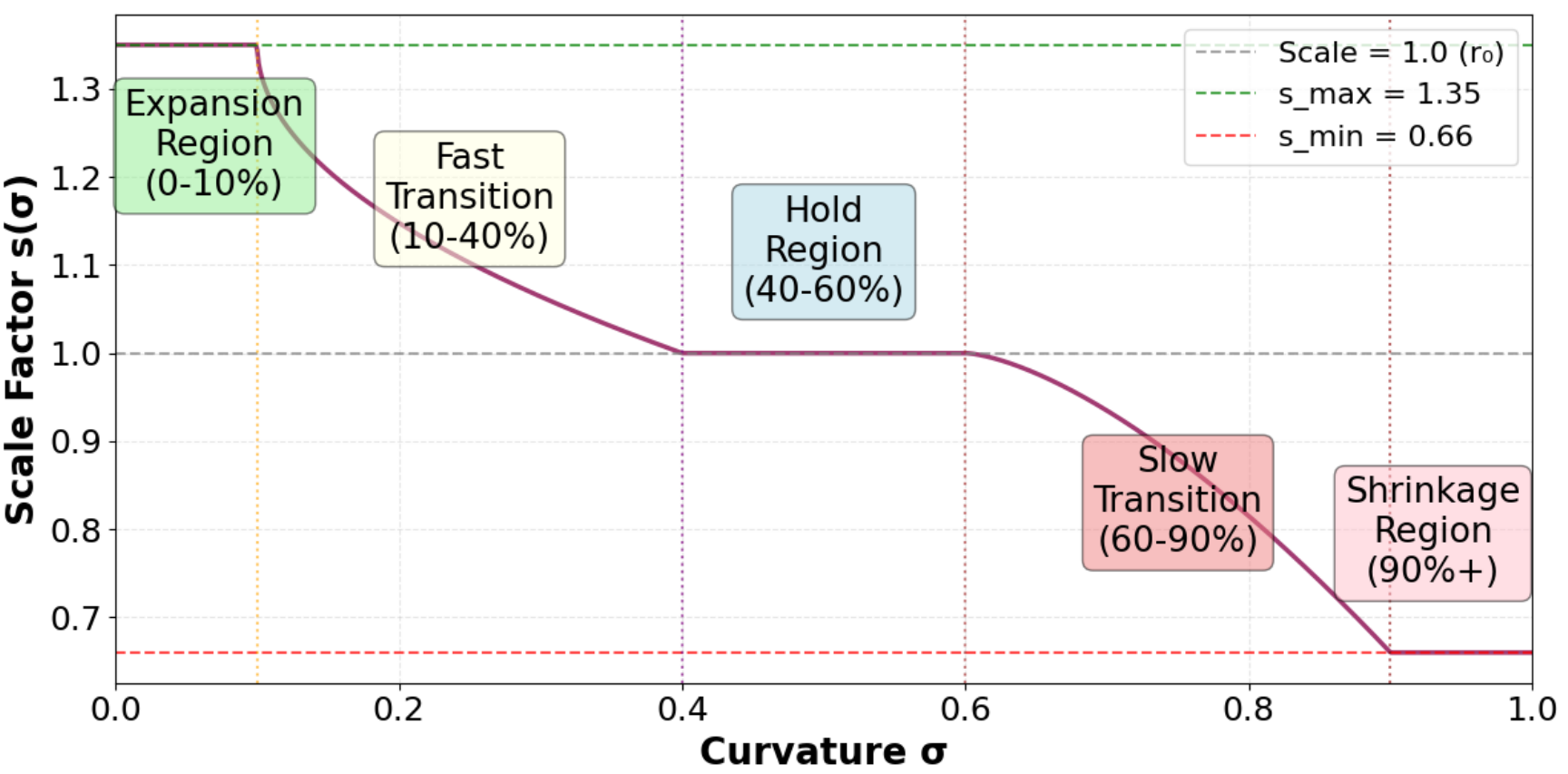}
\caption{Curvature-adaptive radius modulation. There are
several phases in this strategy, separated by vertical lines. For small curvature, the support radius is dilated; it then rapidly relaxes toward the nominal radius and subsequently tapers down as curvature increases.}
\label{fig:radius}
\end{figure}

\subsection{Unsigned distance fields learning}
As an alternative to SDF, UDF-based methods have been proposed for reconstructing surfaces without requiring interior-exterior classification. UDF assigns a non-negative real value to each point in space, representing the shortest distance to the nearest surface, and do not require interior-exterior classification, allowing for the handling of more general shapes.
Neural Unsigned Distance Fields (NDF) \cite{ndf} is one of the earliest and most representative systematic methods for surface reconstruction using UDF.
This study demonstrates that learning UDF with neural networks enables the reconstruction of complex 3D shapes, including non-watertight surfaces and geometrically intricate structures commonly found in real-world data. 
However, many supervised learning approaches tend to overfit to local geometric features and sampling conditions that are prevalent in the training datasets, which often leads to reduced generalization performance.
To address this issue, LoSF-UDF\cite{losf} is guided by the observation that local 3D geometry is often well captured by comparatively simple patterns and is trained on a pseudo-synthetic patch dataset.
Based on this idea, LoSF-UDF trains models using a pseudo-synthetic patch dataset.
Specifically, smooth surfaces and sharp edges are mathematically defined, and the local patches generated from them are used as training data. 
During the reconstruction phase, UDF values are computed based on the point cloud within the patch radius centered on query points arranged on a uniformly spaced 3D grid.
The UDF values obtained for each patch are subsequently used for mesh extraction with DCUDF \cite{dcudf}.
This approach enables general-purpose, high-precision surface reconstruction from point clouds without the need for retraining for specific categories.
However, in LoSF-UDF, the patch radius is fixed, and query points are placed on a uniformly spaced grid, which limits the consideration of local geometric properties.
As a result, reconstruction accuracy may decrease in regions with high curvature or uneven point density.
\begin{table*}[t]
  \centering
  \begin{minipage}{0.63\textwidth}
    \centering
    \caption{Quantitative Results}
    \label{tab:results}
    \begin{tabular}{cc|cccc}
      \hline 
      dataset & Method & CD$\downarrow$ & $F1^{0.005} \uparrow$ & $F1^{0.01} \uparrow$ & NC$\uparrow$ \\ \hline\hline
      \multirow{2}{*}{ShapeNet Cars\cite{cars}} & LoSF-UDF\cite{losf} & 0.794 & 0.506 & 0.865 & 0.594 \\
      & Ours & $\mathbf{0.727}$ & $\mathbf{0.579}$ & $\mathbf{0.899}$ & $\mathbf{0.596}$ \\ \hline
      \multirow{2}{*}{DeepFashion3D\cite{deep}} & LoSF-UDF\cite{losf} & 0.442 & 0.646 & 0.941 & 0.987 \\
      & Ours & $\mathbf{0.433}$ & $\mathbf{0.666}$ & $\mathbf{0.950}$ & 0.987 \\ \hline
    \end{tabular}
  \end{minipage}%
  \hspace{0.05\textwidth}
  \begin{minipage}{0.3\textwidth}
    \centering
    \caption{Runtime (sec)}
    \label{tab:runtime}
    \begin{tabular}{c|cc}
      \hline 
      Method & patch time & udf time \\ \hline\hline
      LoSF-UDF\cite{losf} & 31.025 & 20.198  \\
      Ours & $\mathbf{28.372}$ & $\mathbf{16.367}$ \\ \hline
    \end{tabular}
  \end{minipage}
\end{table*}

\begin{figure}[t]
\centering
\includegraphics[width=\columnwidth]{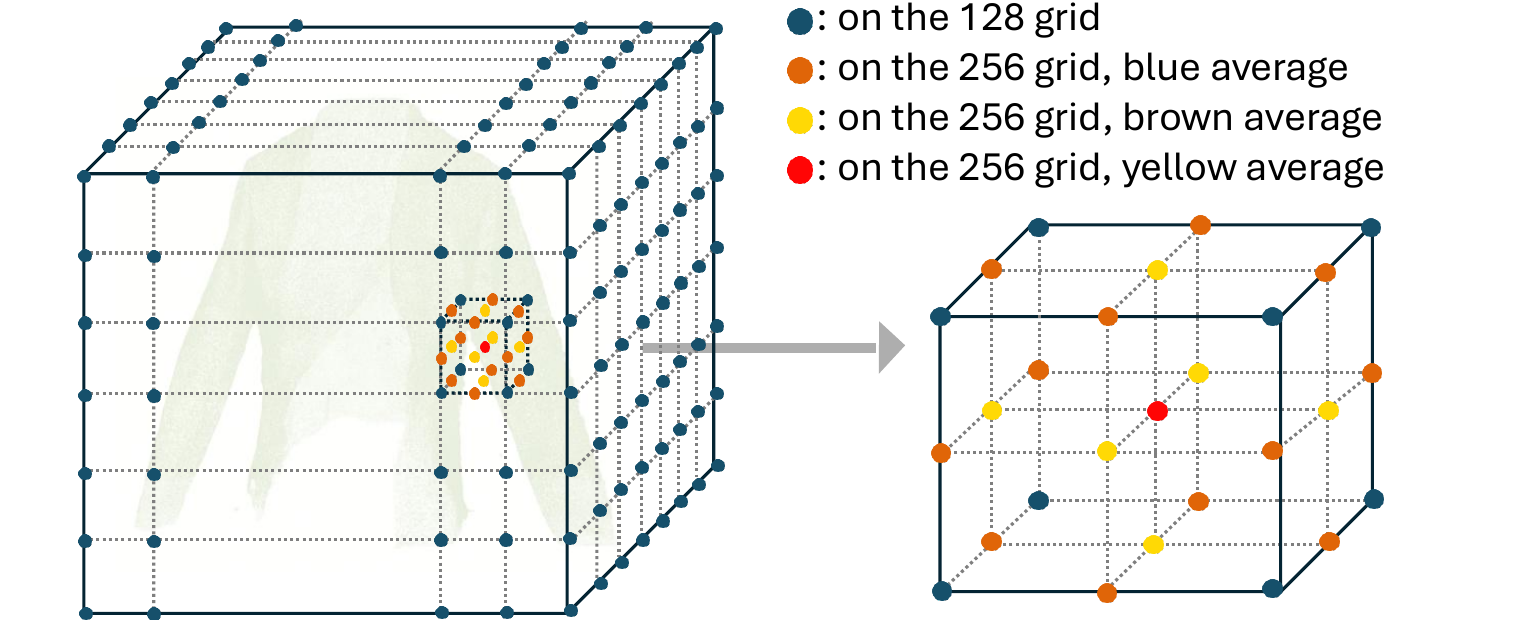}
\caption{Local query augmentation. We introduce two-stage query placement strategy based on curvature. Unevaluated query locations are filled: edge→face→cell centers.}
\label{fig:query}
\end{figure}

\section{Proposed Method}
\subsection{Overview}
We propose a method for point cloud surface reconstruction where patch radius, query point spacing, and resampling strategy are adaptively adjusted based on local curvature.
Query points are initially placed on a $128^3$ grid, with high-curvature areas locally refined to a $256^3$resolution. The patch radius is adjusted according to curvature to maintain geometric details in high-curvature areas and ensure sufficient points in low-curvature regions.
Resampling is performed by duplicating points in high-curvature areas and using centroid-based replication in low-curvature areas. This approach improves reconstruction accuracy while reducing computational complexity.
\subsection{Patch Radius Adjustment}
Our method first places query points $q$ on a uniform grid over the input point cloud and extracts points within a fixed radius $r$ around each query point to obtain initial local regions. Next, the curvature of the point cloud in each region is computed based on surface variation\cite{curve}. The covariance matrix is calculated from the points within the region, and eigenvalues $\lambda_0$, $ \lambda_1$, $\lambda_2$, ($\lambda_0 \le \lambda_1 \le \lambda_2$) are obtained through principal component decomposition. Using these eigenvalues, the curvature $\sigma$ is defined as
\begin{equation}
\label{tab:math}
\sigma_n(\mathbf{p}) = \frac{\lambda_0}{\lambda_0 + \lambda_1 + \lambda_2},
\end{equation}
where 
$n$ represents the region size and \textbf{p} denotes the point cloud within the region. Based on $n$, the radius $r$ is smoothly adjusted according to curvature as shown in Fig. \ref{fig:radius}, and UDF estimation is performed on the point cloud contained within radius $r(\sigma)$ centered at the query point. This design prevents the mixing of regions with different geometric features or separate surfaces/layers in high-curvature regions by reducing the radius, while ensuring sufficient point density in low-curvature regions by increasing the radius. Consequently, this leads to more stable UDF estimation without compromising local geometric characteristics.

\textbf{Radius formulation:} We estimate the 10th, 40th, 60th, and 90th percentiles of the curvature field, 
denoted \(\sigma_{10}, \sigma_{40}, \sigma_{60}, \sigma_{90}\).
and model radius
 $r(\sigma) = r_{0}\, s(\sigma)$ with

\begin{equation}
  \small
  s(\sigma) =
  \begin{cases}
    s_{\max},\!& \!\sigma \le \sigma_{10}, \\[2pt]
    (1-g_{1})\, s_{\max} + g_{1},\!& \!\sigma_{10} < \sigma < \sigma_{40}, \; {\scriptstyle g_{1} = \left(\frac{\sigma - \sigma_{10}}{\sigma_{40} - \sigma_{10}}\right)^{\alpha}}, \\[2pt]
    1,\!& \!\sigma_{40} \le \sigma < \sigma_{60}, \\[2pt]
    1 - (1-s_{\min})\, g_{2},\!& \!\sigma_{60} \le \sigma < \sigma_{90}, \; {\scriptstyle g_{2} = \left(\frac{\sigma - \sigma_{60}}{\sigma_{90} - \sigma_{60}}\right)^{\beta}}, \\[2pt]
    s_{\min},\!& \!\sigma \ge \sigma_{90},
  \end{cases}
\end{equation}
where $s_{\max}=1.35,s_{\min}=2/3,\alpha=0.5,\beta=1.5, r_0=0.018$.

\subsection{Query Point Placement Strategy}
Traditional approaches evaluate queries at all vertices of a $256^3$ uniform grid, delivering accurate reconstructions at the expense of significant computational and memory resources.
In contrast, our approach adopts a two-stage placement strategy that varies grid resolution according to curvature. First, query points are placed on a baseline $128^3$ uniform grid. Subsequently, in regions where curvature $\sigma(n)$ exceeds a threshold, the grid resolution is locally switched to 256, and additional query points are placed. Specifically, $3\times3\times3$ query points on the corresponding 256 grid are added around each existing query point as the center. Any additional points that overlap with existing query points are removed. UDF estimation is then performed for all query points, including the newly added ones. 

For locations on the 256 grid where no query points are placed, UDF values are interpolated using the average of UDF values estimated at surrounding query points as shown in Fig. \ref{fig:query}. First, edge midpoints (brown query in Fig. \ref{fig:query}) are interpolated by averaging the two endpoints, which also exist on the 128 grid with already-estimated UDF values (blue query in Fig. \ref{fig:query}). Next, face centers (yellow query in Fig. \ref{fig:query})are filled by averaging the four edge midpoints adjacent to that face, and finally, cell centers (red query in Fig. \ref{fig:query}) are filled by averaging the six surrounding face centers. This design maintains query point density in high-curvature regions while reducing the number of query points in flat regions, thereby achieving computational cost reduction while preserving fine detail representation.

\subsection{Resampling Strategy}
When a local neighborhood falls short of the target sample count, we normalize differently by curvature.
In smooth regions, we append centroid samples to suppress variance in point distribution and stabilize UDF estimation.
In highly curved regions, we instead duplicate existing samples to avoid biasing the distribution toward the centroid, which would blur sharp features and attenuate fine detail.
This curvature-conditioned policy reduces UDF estimation error in intricate parts while retaining the robustness benefits in flat areas.

\begin{figure*}[t]
\centering
\includegraphics[width=\textwidth]{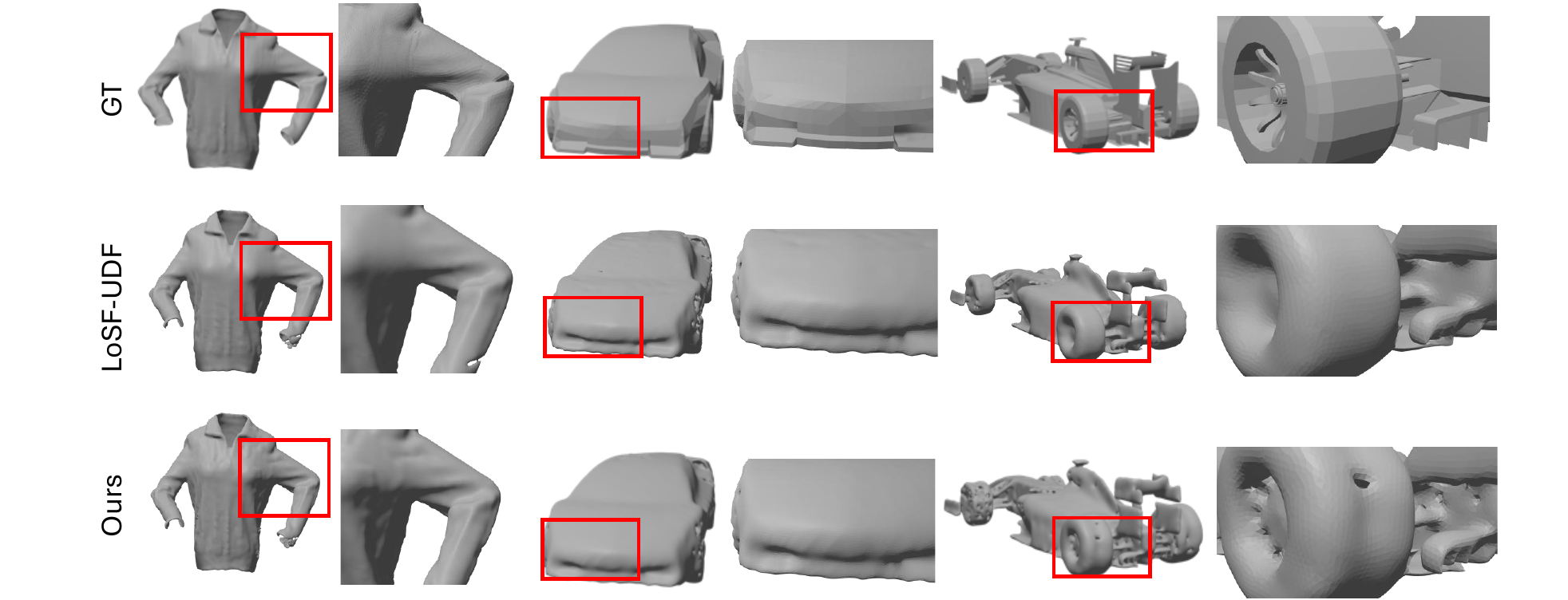}
\caption{Comparison of reconstructed meshes.}
\label{fig:mesh}
\end{figure*}

\section{Experiment}
\subsection{Experimental Details and Evaluation Methods}
We evaluate the point cloud surface reconstruction accuracy of the proposed method. Following the conventional approach, we use the LoSF-UDF proprietary dataset as the training dataset. We used ShapeNet \cite{cars} and DeepFashion3D \cite{deep} as evaluation datasets. ShapeNet consists of synthetic objects, and we randomly extract 100 models from the Car category. DeepFashion3D comprises scanned models of real garments, from which we randomly select 100 models. We employ Chamfer Distance (CD), F1-score, and Normal Consistency (NC) as evaluation metrics.
We re-implement the conventional method as a baseline and conduct reconstruction experiments under identical conditions. Both methods use DCUDF\cite{dcudf} for final mesh extraction.

In this paper, we decompose and report execution time into the following two components.

\textbf{Patch time} : the time required to generate local patches for all query points. For the conventional method, this includes point cloud loading, nearest neighbor search centered at query points, and resampling according to the number of points within the region. For the proposed method, this additionally includes curvature estimation, radius modification based on curvature, and local query point addition in high-curvature regions.

\textbf{UDF time} : the time required to complete UDF estimation for all generated local patches. For the proposed method, this also includes the time for interpolating unevaluated grid points using average values from surrounding points.

These timing measurements were performed on an Intel i7-13700KF CPU and NVIDIA RTX 4070 GPU.

\subsection{Results}
Table 1 reports quantitative results. Across both datasets, our method improves Chamfer Distance (CD) and F1-score over the baseline; Normal Consistency (NC) is comparable or slightly better.
Fig. \ref{fig:mesh} illustrates qualitative results. On planar regions, differences are minimal—indicating that query reduction does not degrade surface fidelity—whereas high-curvature zones (e.g., tire areas) and geometrically intricate underbody parts are recovered more faithfully. Minor localized holes remain, suggesting room for further refinement.
Table 2 summarizes mean runtime. Both patch time and UDF time decrease, yielding an overall reduction in computational cost. This stems from the two-stage strategy that starts at a 128³ lattice and promotes only high-curvature neighborhoods to 256³, systematically reducing the number of evaluated queries relative to uniform 256³ processing. Although UDF time in our method additionally includes imputing values at unevaluated lattice sites via neighborhood averaging, the savings from fewer UDF evaluations dominate, resulting in a net speedup.

\section{Conclusion}

We propose a point cloud surface reconstruction approach that adapts local regions according to curvature. The results indicate higher reconstruction fidelity together with reduced computational cost. Future research should focus on mitigating newly observed local defects.

\section{References}

\vspace{-2em}
\small
\setlength\itemsep{-0.2em}
\setlength\baselineskip{8pt}

\end{document}